\documentclass[10pt,twocolumn,letterpaper]{article}

\usepackage{iccv}
\usepackage{times}
\usepackage{epsfig}
\usepackage{graphicx}
\usepackage{amsmath}
\usepackage{amssymb}

\usepackage[pagebackref=true,breaklinks=true,letterpaper=true,colorlinks,bookmarks=false]{hyperref}

\usepackage{booktabs}

\usepackage{graphicx}

\usepackage{bm}
\usepackage{algorithm}
\usepackage{algorithmic}
\usepackage{multirow}
\usepackage{color}

\usepackage{graphicx}
\usepackage{float} 
\usepackage{subcaption}

\iccvfinalcopy 

\def\iccvPaperID{****} 

\ificcvfinal\fi

\begin{document}

\title{Membrane Potential Batch Normalization for Spiking Neural Networks}

\author{Yufei Guo\thanks{Equal contribution.}, Yuhan Zhang\footnote[1]{}, Yuanpei Chen, Weihang Peng, Xiaode Liu, Liwen Zhang,\\
Xuhui Huang, Zhe Ma\thanks{Corresponding author.}\\
 Intelligent Science \& Technology Academy of CASIC, China\\
 Scientific Research Laboratory of Aerospace Intelligent Systems and Technology, China\\
{\tt\small yfguo@pku.edu.cn, yuhanzhang@pku.edu.cn, mazhe\_thu@163.com}
}

\maketitle
\ificcvfinal\thispagestyle{empty}\fi

\begin{abstract}
As one of the energy-efficient alternatives of conventional neural networks (CNNs), spiking neural networks (SNNs) have gained more and more interest recently. 
To train the deep models, some effective batch normalization (BN) techniques are proposed in SNNs.
All these BNs are suggested to be used after the convolution layer as
usually doing in CNNs.
However, the spiking neuron is much more complex with the spatio-temporal dynamics. 
The regulated data flow after the BN layer will be disturbed again by the membrane potential updating operation before the firing function, i.e., the nonlinear activation. 
Therefore, we advocate adding another BN layer before the firing function to normalize the membrane potential again, called MPBN.
To eliminate the induced time cost of MPBN, we also propose a training-inference-decoupled re-parameterization technique to fold the trained MPBN into the firing threshold. 
With the re-parameterization technique, the MPBN will not introduce any extra time burden in the inference.
Furthermore, the MPBN can also adopt the element-wised form, while these BNs after the convolution layer can only use the channel-wised form.
Experimental results show that the proposed MPBN performs well on both popular non-spiking static and neuromorphic datasets. Our code is open-sourced at \href{https://github.com/yfguo91/MPBN}{MPBN}.

\end{abstract}

\section{Introduction}
\label{sec:intro}

Emerged as a biology-inspired method, spiking neural networks (SNNs) have received much attention in artificial intelligence and neuroscience recently~\cite{guo2023direct,2009Spiking,xu2023constructing,xu2023biologically,shen2023esl,xu2022hierarchical,xu2021robust}.
SNNs deal with binary event-driven spikes as their activations and therefore the multiplications of activations and weights can be substituted for additions or only keeping silents. 
Benefitting from such a computation paradigm, SNNs derive extreme energy efficiency and run efficiently when
implemented on neuromorphic hardware~\cite{2015TrueNorth,2019Towards,2018Loihi}.

Despite the SNN has achieved great success in diverse fields including pattern recognition~\cite{2020Incorporating,guo2022real,guo2023joint,guo2023neuroclip}, object detection~\cite{kim2020spiking}, language processing~\cite{xiao2022towards}, robotics~\cite{dewolf2021spiking}, and so on, its development is deeply  inspired by the experience of convolutional neural networks (CNNs) in many aspects. However, the spiking neuron model along with the rich spatio-temporal dynamic makes SNNs much different from CNNs, and directly transferring some experience of CNNs to SNNs without any modifications may be not a good idea.
As one of the famous techniques in CNNs, the batch normalization(BN) technique shows great advantages. It can reduce the gradient exploding/vanishing problem, flatten the loss landscape, and reduce the internal covariate shift, thus being widely used in CNNs. 
There are also some works trying to apply normalization approaches in the SNN field to help model convergence.
For example, inspired by BN in CNNs, NeuNorm~\cite{2018Direct} was proposed to normalize the data along the channel dimension. Considering that the temporal dimension is also  important in SNNs, threshold-dependent batch normalization 
(tdBN)~\cite{2020Going} then extended the scope of BN to the additional temporal dimension. Subsequently, to better depict the differences of data flow distributions in different time dimensions, the temporal batch normalization through time (BNTT)~\cite{2020Revisiting}, postsynaptic potential normalization (PSP-BN)~\cite{Rethinking2022}, and temporal effective batch normalization (TEBN)~\cite{duan2022temporal} that regulate the data flows with multiple BNs on different time steps were proposed.

\begin{figure}[t]
	\centering
	\includegraphics[width=0.40\textwidth]{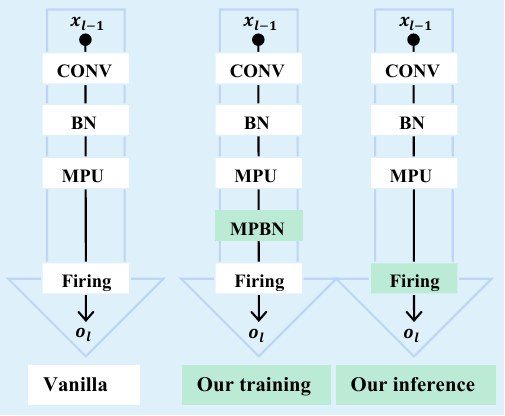} 
	\caption{The difference between our SNN with MPBN and the vanilla SNN. We add another BN layer after membrane potential updating (MPU) operation in the training. The MPBN can be folded into the firing threshold and then the homogenous firing threshold will be transformed into different ones.}
	\label{titlefig}
\end{figure}

However, all these BNs proposed in SNNs are advised to be used after convolution layers as usually doing in CNNs. This ignores the fact that the nonlinear transformation in the SNN spiking neuron is much more complex than that of the ReLU neuron. In the spiking neuron, the data flow after the convolution layer will be first injected into the residual membrane potential (MP) coming from the previous time step to generate a new MP at the current time step. And then the neuron will fire a spike or keep silent still based on whether or not the new MP is up to the firing threshold. Obviously, though the data flow has been normalized by the BN after the convolution layer, it will be disturbed again by the residual MP in the membrane potential updating process. Therefore, we advocate also adding a BN layer after MP updating to regulate the data flow once again, called MPBN. Furthermore, we also propose a training-inference-decoupled re-parameterization technique in SNNs to fold the trained MPBN into the firing threshold. Hence, the MPBN will not induce any extra burden in the inference but a trivial burden in the training. The MPBN can be extended to channel-wised MPBN and element-wised MPBN further, which is very different from that of CNNs where only channel-wised normalization can be folded into weights. 
The difference between our SNN with MPBN and the vanilla SNN is illustrated in Fig.~\ref{titlefig}. Our main contributions are as follows:

\begin{itemize}
	
\item We propose to add another BN layer after the membrane potential updating operation named MPBN to handle the data flow disturbance in the spiking neuron.
The experiment shows that MPBN can flatten the loss landscape further, thus benefiting model convergence and task accuracy. 

\item We also propose a re-parameterization method to decouple the training-time SNN and the inference-time SNN. In specific, we propose a method to fold the trained MPBN parameter into the firing threshold. Therefore, MPBN can be seen as only training auxiliary manner free from burdens in the inference-time. This re-parameterization method is suitable for both channel-wised MPBN and element-wised MPBN.

\item Extensive experiment results show that the SNN trained with the MPBN is highly effective compared with other state-of-the-art SNN models on both static and dynamic datasets, e.g., 96.47\% top-1 accuracy and 79.51\% top-1 accuracy are achieved on the CIFAR-10 and CIFAR-100 with only 2 time steps. 

\end{itemize}


\section{Related Work}

\subsection{Learning of Spiking Neural Networks}

There are three kinds of learning algorithms of SNNs, including unsupervised learning~\cite{Peter2015Unsupervised,2018ABiologically}, converting ANN to SNN (ANN2SNN)~\cite{2019Going,hao2023reducing,hao2023bridging},  and supervised learning~\cite{guo2023joint,li2021differentiable,Guo_2022_CVPR}. 
Unsupervised learning adopts some biological mechanism to update the SNN model, i.e., the spike-timing-dependent plasticity (STDP) approach~\cite{2020Spatial},  thus being considered a biologically plausible method. 
However,  STDP cannot help train large-scale networks yet, thus it is usually limited to small datasets and non-ideal performance.
The ANN-SNN conversion approach~\cite{2020Deep,li2021free} obtains an SNN by reusing well-trained homogeneous ANN parameters and replacing the ReLU neuron with a spiking neuron. 
Since the ANN model is easier to train and reach high performance, the ANN-SNN conversion method provides an interesting way to generate an SNN in a short time with competitive performance.
However, the converted SNN will lose the rich temporal dynamic behaviors and thus cannot handle neuromorphic datasets well. 
Supervised learning~\cite{2021Deep,2018Spatio,guo2023rmploss} adopts the surrogate gradient (SG) approach to train SNNs with error backpropagation. 
It can handle temporal data and provide decent performance with few time steps on the large-scale dataset, thus having received much attention recently.
For a more detailed introduction, please refer to the recent SNN survey~\cite{guo2023direct}.
Our work falls under the supervised learning.

\subsection{Normalization in Spiking Neural Networks}

The batch normalization technique was originally introduced as a kind of  training auxiliary method by~\cite{ioffe2015batch} in CNNs. It uses the weight-summed input over a mini-batch of training cases to compute a mean and variance and then uses them to regulate the summed input. This simple operation can derive many benefits. i) It reduces the internal covariate shift (ICS), thus accelerating the training of a deep neural network. ii) It makes the network insensitive to the scale of the gradients, thus a higher learning rate can be chosen to accelerate the training. iii) It makes the network suitable for more nonlinearities by preventing the network from getting stuck in the saturated modes. With these advantages, more kinds of BNs were proposed, including  layer normalization~\cite{ba2016layer}, group normalization~\cite{2018Group}, instance normalization~\cite{ulyanov2017instance}, and switchable normalization~\cite{luo2019differentiable}. 

There are also some works that modify and apply normalization approaches in the SNN field. For example, NeuNorm~\cite{2018Direct} also normalizes the feature map along the channel dimension like BN in CNNs. Recently, some methods were proposed to normalize the feature map from both the channel dimension and temporal dimension to take care of the spatio-temporal characteristics of the SNN, such as the threshold-dependent batch normalization (tdBN)~\cite{2020Going}. It extends the scope of BN to the additional temporal dimension by adopting a 3DBN-like normalization method in CNNs. Note that, the tdBN can be folded into the weights, thus inducing no burden in the inference time. Nevertheless, NeuNorm and tdBN still use the shared parameters along the temporal dimension. Some works argued that the distributions of data in different time steps vary wildly and that using shared parameters is not a good choice. Subsequently, the temporal batch normalization through time (BNTT)~\cite{2020Revisiting}, postsynaptic potential normalization (PSP-BN)~\cite{Rethinking2022}, and temporal effective batch normalization (TEBN)~\cite{duan2022temporal} were proposed. These BNs regulate the data flow utilizing different parameters through time steps.
Though these BNs with different BN parameters on different time steps can train more well-performed SNN models, their parameters can not be folded into the weights, thus will increase the computations and running time in the inference.

Nevertheless, all these BNs in the SNN field are advised to be used after convolution layers. 
However, the data flow after the convolution layer will not be presented to the firing function directly but to the membrane potential updating function first. 
Hence, the data flow will be disturbed again before reaching the firing function.
To this end, in this paper, we add another BN after the membrane potential updating function, called the MPBN to retain normalized data flow before the firing function.

\section{Preliminary}

\subsection{Leaky Integrate-and-Fire Model}

Different from CNNs, SNNs use binary spikes to transmit information. 
In the paper, we use the widely used Leaky-Integrate-and-Fire (LIF) neuron model~\cite{nahmias2013leaky} to introduce the unique spatial-temporal dynamic of the spiking model.
First, we introduce the notation rules used here as follows. 
Vectors or tensors are denoted by bold italic letters, i.e., $\bm{x}$ and $\bm{o}$ represent the input and output variables respectively. 
Matrices are denoted by bold capital letters. For instance, $\mathbf{W}$ is the weight matrix. 
The constant is denoted by small letters. 

In LIF, the membrane potential is updated by
\begin{equation}
    \bm{u}^{(t+1), \text{pre}} = \tau\bm{u}^{(t)} + \bm{c}^{(t+1)}, \text{where } \bm{c}^{(t+1)} = \mathbf{W} \bm{x}^{(t+1)}, \label{eq_lif}
\end{equation}
where $\bm{u}$ represents the membrane potential and $\bm{u}^{(t+1), \text{pre}}$ is the updated membrane potential at time step $t+1$, $\bm{c}^{(t+1)}$ is the pre-synaptic input at time step $t+1$, which is charged by weight-summed input spikes $\bm{x}^{(t+1)}$, and $\tau$ is a constant within $(0, 1)$, which controls the leakage of the membrane potential.
Then, when the updated membrane potential $\bm{u}^{(t+1), \text{pre}}$ is up to the firing threshold $V_{\rm th}$, the LIF spiking neuron will fire a spike as bellow,
\begin{equation}
    \begin{split}
    \bm{o}^{(t+1)} = 
    \begin{cases}
        1 & \text{if } \bm{u}^{(t+1), \text{pre}} > V_{\rm th} \\
        0 & \text{otherwise}
    \end{cases}, \\
    \bm{u}^{(t+1)} = \bm{u}^{(t+1), \text{pre}}\cdot(1 - \bm{o}^{(t+1)}). \label{eq_fire}
    \end{split}
\end{equation}
After firing, the spike output $\bm{o}^{(t+1)}$ at time step $t+1$ will be transmitted to the next layer and become its input. At the same time, the updated membrane potential will be reset to zero and becomes $\bm{u}^{(t+1)}$ to join the neuron processing at the next time step.

\textbf{The Classifier in the SNN. }
In a classification model, the final output is used to compute the $\mathrm{Softmax}$ and predict the desired class object.
In an SNN model, if we also use LIF neurons at the output layer to fire spikes and use the number of spikes to compute the probability, too much information will be lost. 
Therefore, we only integrate the output and do not fire them across time, as doing in recent work~\cite{Guo_2022_CVPR,guo2022real,2020Incorporating}. 
\begin{equation}
    \bm{o}_{\text{out}} = \frac{1}{T}\sum_{t=1}^{T} \bm{c}_{\text{out}}^{(t)} =  \frac{1}{T}\sum_{t=1}^{T}\mathbf{W} \bm{x}^{(t)}.
    \label{eq_out}
\end{equation}
Then, the cross-entropy loss is computed based on the true label and $\mathrm{Softmax}(\bm{o}_{\text{out}})$.

\subsection{Batch Normalization in SNNs}

Batch normalization can effectively reduce the internal covariate shift and alleviate the gradient vanishing or explosion problem for training networks, thus having been widely used in CNNs. Fortunately, BN can also be used in SNNs. Considering a spiking neuron with input $\bm{c} = \{ \bm{c}^{(1)}, \bm{c}^{(2)}, \dots,\bm{c}^{(t)}, \dots \}$, where $t$ is the time step, BN regulate the input at each time step as follows,
\begin{equation}\label{bn}
	\tilde{\bm{c}}_i^{(t)} =  \frac{\bm{c}_i^{(t)}- \bm{\mu}_i}{\sqrt{\bm{\sigma}_i^2+\epsilon}},
\end{equation}
where $\bm{c}_i^{(t)}$ is the input in $i$-th channel at $t$-th time step, $\bm{\mu}_i$ and $\bm{\sigma}_i$ are the mean and variance of input in channel dimension, and $\epsilon$ is a small constant to avoid denominator being zero. To ensure BN can represent the identity transformation, the normalized vector $\tilde{\bm{c}}_i^{(t)}$ is scaled and shifted in a learning manner as follows,
\begin{equation}\label{bn2}
	{\rm{BN}}({\bm{c}}_i^{(t)} )= \bm{\lambda}_i \tilde{\bm{c}}_i^{(t)} + \bm{\beta}_i,
\end{equation}
where $\bm{\lambda}_i$ and $\bm{\beta}_i$ are channel-wised learnable parameters.

\section{Methodology}

This section first introduces the specific form of membrane potential batch normalization. Then the re-parameterization technique that how to fold the MPBN into $V_{\rm th}$ will be introduced in detail. Next, some key details for training the SNN  and the pseudocode for the training and inference of our SNN will be given. Finally, we will provide plenty of  ablation studies and the comparison of the loss landscape of the models with or without MPBN to show the effectiveness of the proposed method.

\subsection{Membrane Potential Batch Normalization}

As abovementioned, we argue that though the data flow has been normalized by the BN after the convolution layer, it will be disturbed again by the membrane potential updating operation. To better depict this, we give the vanilla form of LIF neuron with BN first as follows,
\begin{equation}
    \bm{u}^{(t+1), \text{pre}} = \tau\bm{u}^{(t)} + {\rm BN}(\mathbf{W} \bm{x}^{(t+1)}), \label{eq_lif_bn}
\end{equation}
where $\tau$ is 0.25 in the paper following~\cite{guo2022real,li2021differentiable,2020LISNN}. To regulate the disturbed data flow once again, We further embed another BN after the membrane potential updating operation, called MPBN. The LIF neuron with MPBN can be updated as
\begin{equation}
    \tilde{\bm{u}}^{(t+1), \text{pre}} = {\rm MPBN}(\bm{u}^{(t+1), \text{pre}}).
\end{equation}
Obviously, $\bm{u}^{(t+1), \text{pre}}$ will be scaled and sifted, and some $\bm{u}^{(t+1), \text{pre}}$ less than $V_{\rm th}$ may be greater than $V_{\rm th}$ with MPBN and vice versa. This is abhorrent with the biology and MPBN will cause some extra computation burden in the inference compared with the vanilla one. To solve this problem, we also propose a training-inference-decoupled re-parameterization technique here.

\begin{algorithm}[t]
	\caption{Training and inference of our SNN.}
	\label{alg:algorithm}
    \textcolor{blue}{\textbf{Training}}\\
	\textbf{Input}: An SNN to be trained with MPBN; training dataset; total training iteration: $I_{\rm train}$.\\
	\textbf{Output}: The well-trained SNN.
	\begin{algorithmic}[1] 
		\FOR {all $i = 1, 2, \dots , I_{\rm train}$ iteration}
		\STATE Get mini-batch training data, $\bm{x}_{\rm in}(i)$ and class label, $\bm{y}(i)$;
		\STATE Feed the  $\bm{x}_{\rm in}(i)$ into the SNN and regulate the data flow by BN and MPBN;
		\STATE Calculate the SNN output, $\bm{o}_{\rm out}(i)$ by Eq.~\ref {eq_out} ;
		\STATE Compute classification loss $L_{\rm CE}={\mathcal{L}_{\rm CE}}(\bm{o}_{\rm out}(i),\bm{y}(i))$;
		\STATE Calculate the gradient w.r.t. $\mathbf{W}$ by Eq.~\ref {eq:gradiet};
		\STATE Update $\mathbf{W}$: $(\mathbf{W} \leftarrow  \mathbf{W} - \eta \frac{\partial {L}}{\partial {\mathbf{W}}})$ where $\eta$ is learning rate.
		\ENDFOR\\
	\end{algorithmic}
    \textcolor{blue}{\textbf{Re-parameterization}}\\
    \textbf{Input}: The trained SNN with MPBN; total number of MPBN: $n$.\\
    \textbf{Output}: The re-parameterized trained SNN without MPBN but diverse $V_{\rm th}$.
    \begin{algorithmic}[1] 
		\FOR {all $i = 1, 2, \dots , n$ number}
		\STATE Fold the parameters of $i$-th MPBN into $i$-th $V_{\rm th}$ by Eq.~\ref {eq_lif_bn2};
		\ENDFOR\\
	\end{algorithmic}
    \textcolor{blue}{\textbf{Inference}}\\
	\textbf{Input}: The re-parameterized trained SNN; test dataset; total test iteration: $I_{\rm test}$.\\
	\textbf{Output}: The output.
	\begin{algorithmic}[1] 
		\FOR {all $i = 1, 2, \dots , I_{\rm test}$ iteration}
		\STATE Get mini-batch test data, $\bm{x}_{\rm in}(i)$ and class label, $\bm{y}(i)$ in test dataset;
        \STATE Feed the  $\bm{x}_{\rm in}(i)$ into the reparameterized SNN without MPBN;
		\STATE Calculate the SNN output, $\bm{o}_{\rm out}(i)$ by Eq.~\ref {eq_out} ;
		\STATE   Compare the classification factor $\bm{o}_{\rm out}(i)$ and $\bm{y}(i)$ for classification.
		\ENDFOR\\
	\end{algorithmic}
\end{algorithm}

\subsection{Re-parameterization}
With MPBN, the firing function will be updated as
\begin{equation}
    \bm{o}^{(t+1)} = 
    \begin{cases}
        1 & \text{if } {\rm MPBN}(\bm{u}^{(t+1), \text{pre}}) > V_{\rm th} \\
        0 & \text{otherwise}
    \end{cases}. \label{eq_fire_bn}
\end{equation}
If we unfold the MPBN, the above equation will be re-organized as
\begin{equation}
    \bm{o}_i^{(t+1)} = 
    \begin{cases}
        1 & \text{if } \bm{\lambda}_i \frac{\bm{u}_i^{(t+1), \text{pre}} - \bm{\mu}_i}{\sqrt{\bm{\sigma}_i^2}} + \bm{\beta}_i > V_{\rm th} \\
        0 & \text{otherwise}
    \end{cases}. \label{eq_fire_bn2}
\end{equation}
By folding the MPBN to $V_{\rm th}$, the firing function will be further updated as
\begin{equation}
\begin{split}
\bm{o}_i^{(t+1)} = 
    \begin{cases}
        1 & \text{if } \bm{u}^{(t+1), \text{pre}} > (\bm{\tilde{V}}_{\rm th})_i \\
        0 & \text{otherwise}
    \end{cases},\\
    \text{where } (\bm{\tilde{V}}_{\rm th})_i = \frac{(V_{\rm th} - \bm{\beta}_i){\sqrt{\bm{\sigma}_i^2}}}{\bm{\lambda}_i} + \bm{\mu}_i.
    \label{eq_lif_bn2}
\end{split}
\end{equation}
It can be seen that by absorbing some parameters from MPBN, $V_{\rm th}$ will be transformed to another channel-wised $(\bm{\tilde{V}}_{\rm th})_i$. In this way, the extra computation burden caused by MPBN will be eliminated again in the inference time. Furthermore, the diversity of the spiking neuron will be improved with abundant firing parameters as the learnable firing threshold in other work~\cite{2018Long,wang2022ltmd}.

\subsection{Training Framework}

In the paper, the spatial-temporal backpropagation (STBP) algorithm~\cite{2018Direct} is adopted to train the SNN models. STBP treats the SNN model as a self-recurrent neural network thus enabling an error backpropagation mechanism following the same principles as in CNNs. However, there is still a problem impeding the direct training of SNNs.  
To demonstrate this problem, we formulate the gradient at the layer $l$ by the chain rule, given by
\begin{equation}\label{eq:gradiet}
 \frac{\partial {L}}{\partial {\mathbf{W}^l}} = \sum_t (\frac{\partial {L}}{\partial {\bm{o}^{(t),l}}} \frac{\partial {\bm{o}^{(t),l}}}{\partial {\bm{u}^{(t),l}}} +  \frac{\partial {L}}{\partial {{\bm{u}^{(t+1),l}}}} \frac{\partial {{\bm{u}^{(t+1),l}}}}{\partial {{\bm{u}^{(t),l}}}} )\frac{\partial {{\bm{u}^{(t),l}}}}{\partial {\mathbf{W}^l}},
\end{equation}
where $\frac{\partial {\bm{o}^{(t),l}}}{\partial {\bm{u}^{(t),l}}}$ is the gradient of firing function at at $t$-th time step in $l$-th layer.
Obviously, the non-differentiable firing activity of the spiking neuron will result in zero gradients everywhere, while infinity at $V_{\rm th}$. Therefore, the gradient descent $(\mathbf{W}^l \leftarrow  \mathbf{W}^l - \eta \frac{\partial {L}}{\partial {\mathbf{W}^l}})$ either freezes or updates to infinity in the backpropagation.
To handle this problem, here,  we also adopt the commonly used STE surrogate gradients as doing in other surrogate gradients (SG)  methods~\cite{2020DIET,Guo_2022_CVPR}. Mathematically, it is defined as:
\begin{eqnarray}\label{sg}
	\frac{d{\bm{o}}}{d{\bm{u}}}=
	\left\{
	\begin{array}{lll}
		1, \ \ \text{if} \; 0 \le {\bm{u}} \le 1 \\
		0, \ \ \text{otherwise}
	\end{array}
	\right..
\end{eqnarray}
Then, the SNN model can be trained end-to-end. 
The training and inference of our SNN  are detailed in
Algo.~\ref{alg:algorithm}.

\begin{figure}[t]
\centering
    \begin{subfigure}[t]{0.23\textwidth}
       \centering
        \includegraphics[width=0.99\textwidth]{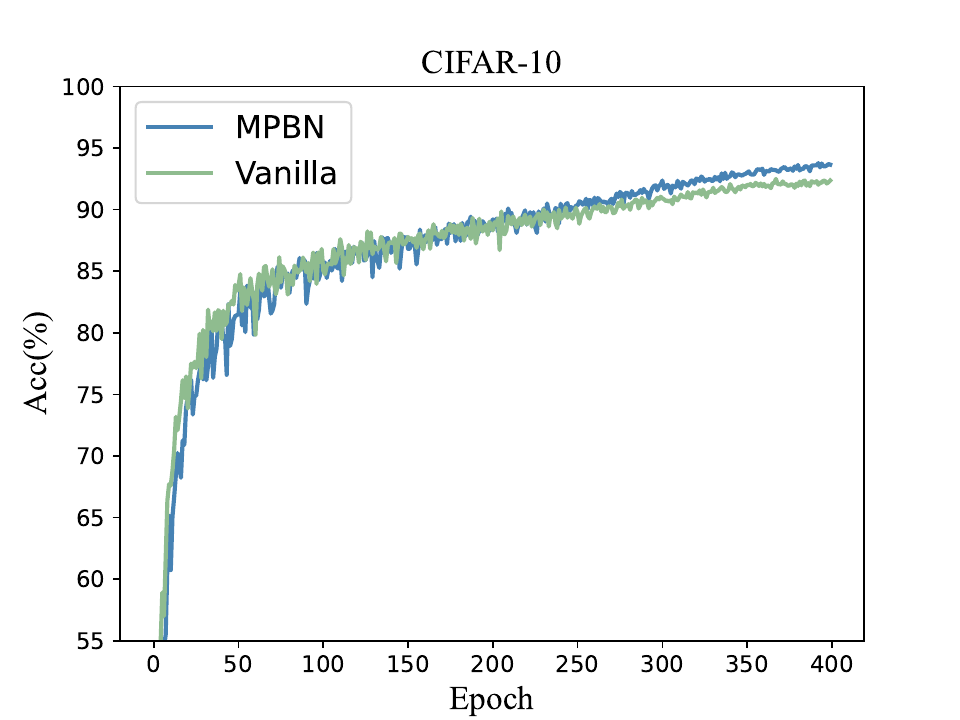}\caption{CIFAR-10}
        \label{fig:Case1}
    \end{subfigure}
    \begin{subfigure}[t]{0.23\textwidth}
       \centering
        \includegraphics[width=0.99\textwidth]{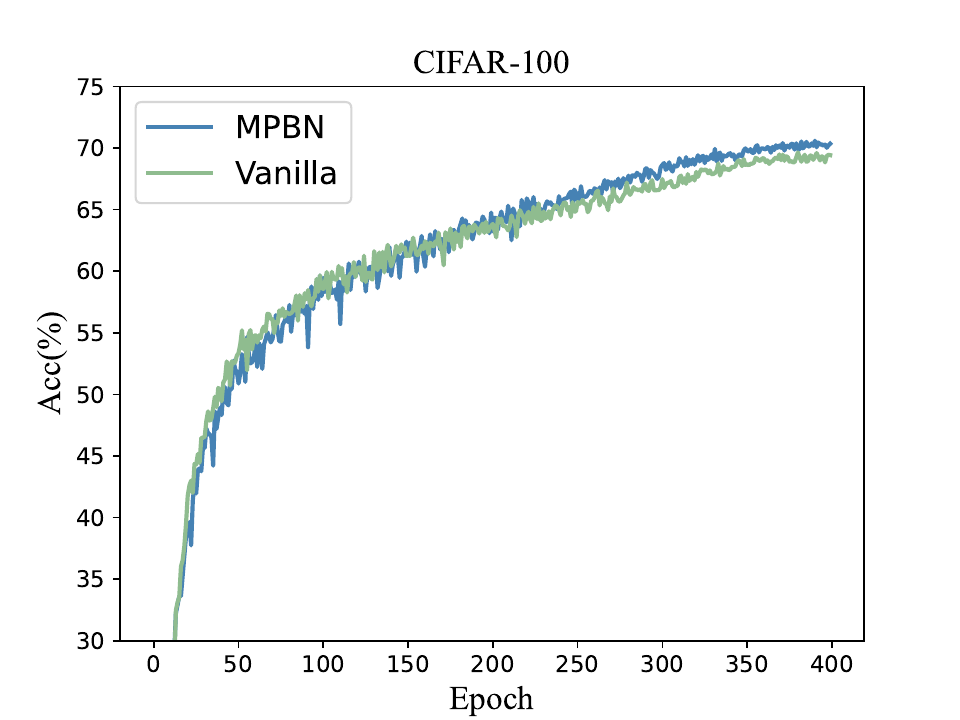}\caption{CIFAR-100}
        \label{fig:Case1}
    \end{subfigure}
	\caption{The accuracy curves of spiking ResNet20 with or without MPBN using 2 time steps on CIFAR-10 (left) and CIFAR-100 (right). The MPBN based SNNs obviously enjoy higher accuracy and easier convergence.}
	\label{abla}
\end{figure}
\subsection{Ablation Study}

To verify the effectiveness of the MPBN, a lot of ablative studies using spiking ResNet20 architecture along with different time steps were conducted on the CIFAR-10 and CIFAR-100 datasets. The results of top-1 accuracy of these models are shown in Tab.~\ref {ablampbn}. It's can be seen that the test accuracy of the SNNs with MPBN is always higher than these vanilla counterparts. For example, the accuracy of baseline SNN with 1 time step is 90.40\%, while with MPBN,  it will increase up to 92.22\%, which is a huge improvement (more than 2.0\%) in the SNN field.
Moreover, we also show the test accuracy curves of ResNet20 using MPBN and its vanilla counterpart with 2 time steps on CIFAR-10/100 during training in Fig.~\ref {abla}. It can be observed obviously that the SNNs with MPBN  also perform better on convergence speed. To sum up, the proposed MPBN can both improve accuracy and convergence speed, which are very important aspects in deep learning.
\begin{table}[]	
	\centering	
	\caption{Ablation experiments for MPBN.}	
	\label{ablampbn}	
	\begin{tabular}{clcc}	
		\toprule
		Dataset & Method & Time step & Accuracy \\	
		\toprule
		\multirow{6}{*}{CIFAR-10}	
		& baseline & 1 & 90.40\%   \\
		& w/ MPBN  & 1 & 92.22\%   \\
        \cline{2-4}
		& baseline & 2 & 92.80\%   \\
		& w/ MPBN  & 2 & 93.54\%   \\	
	  \cline{2-4}	
        & baseline & 4 & 93.85\%   \\
		& w/ MPBN  & 4 & 94.28\%   \\
        \cline{1-4}
        \multirow{6}{*}{CIFAR-100}	
		& baseline & 1 & 67.94\%   \\
		& w/ MPBN  & 1 & 68.36\%   \\
        \cline{2-4}
		& baseline & 2 & 70.18\%   \\
		& w/ MPBN  & 2 & 70.79\%   \\	
	  \cline{2-4}	
        & baseline & 4 & 71.77\%   \\
		& w/ MPBN  & 4 & 72.30\%   \\
		\bottomrule 	
	\end{tabular}	
\end{table}
\begin{figure}[t]
\centering
    \begin{subfigure}[t]{0.23\textwidth}
       \centering
        \includegraphics[width=0.99\textwidth]{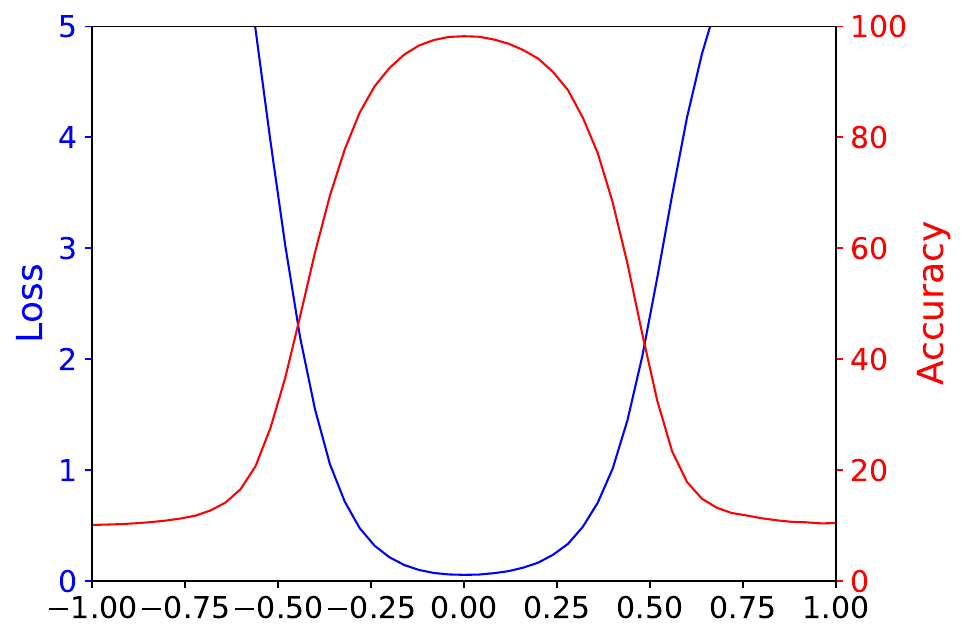}\caption{MPBN}
        \label{fig:Case1}
    \end{subfigure}
    \begin{subfigure}[t]{0.23\textwidth}
       \centering
        \includegraphics[width=0.99\textwidth]{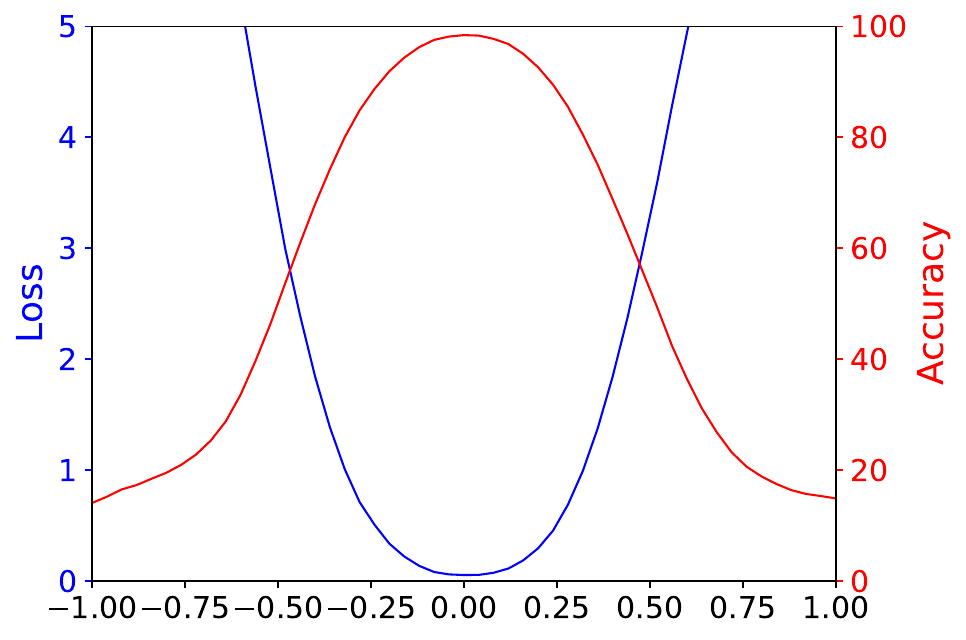}\caption{Vanilla}
        \label{fig:Case1}
    \end{subfigure}
	\caption{The 1D loss landscape of spiking ResNet20 with and without MPBN.}
	\label{loss}
\end{figure}
\begin{table*}[!h]
	\centering	
	\caption{Comparison with SoTA methods on CIFAR-10.}	
	\label{tab:Comparisoncifar}	
	\begin{tabular}{lllccc}	
		\toprule
		Dataset & Method & Type & Architecture & Timestep & Accuracy \\	
		\toprule
		\multirow{34}{*}{CIFAR-10}	
		& SpikeNorm~\cite{2019Going} & ANN2SNN & VGG16 & 2500 & 91.55\%   \\	
		& Hybrid-Train~\cite{2020Enabling} & Hybrid training & VGG16 & 200 & 92.02\%   \\
		& Spike-basedBP~\cite{2019Enabling} & SNN training & ResNet11 & 100 & 90.95\%   \\
		& Joint A-SNN~\cite{guo2023joint} & SNN training & ResNet18 & 4 & 95.45\%   \\  	
		& GLIF~\cite{yao2022glif} & SNN training & ResNet19 & 2 & 94.44\%   \\ 
		& PLIF~\cite{2020Incorporating} & SNN training & PLIFNet & 8 & 93.50\%   \\
		\cline{2-6}
		& \multirow{4}{*}{Diet-SNN~\cite{2020DIET}} & \multirow{4}{*}{SNN training} & \multirow{2}{*}{VGG16} & 5 & 92.70\%   \\ 
		& &  &  & 10 & 93.44\%   \\ 
		\cline{4-6}
		&  &  & \multirow{2}{*}{ResNet20} & 5 & 91.78\%   \\ 
		&  &  &  & 10 & 92.54\%   \\  
		\cline{2-6}
		& \multirow{3}{*}{RecDis-SNN~\cite{Guo_2022_CVPR}} & \multirow{3}{*}{SNN training} & \multirow{3}{*}{ResNet19} 
		& 2 & 93.64\%   \\
		&  &  &											                                  & 4 &  95.53\%   \\
		&  &  &											                                   & 6 &  95.55\%   \\
		\cline{2-6}
		& \multirow{3}{*}{Dspike~\cite{li2021differentiable}} & \multirow{3}{*}{SNN training} & \multirow{3}{*}{ResNet20} 
		& 2 & 93.13\%   \\
		&  &  &											                                  & 4 & 93.66\%   \\
		&  &  &											                                   & 6 & 94.25\%   \\
		\cline{2-6}
		& \multirow{3}{*}{STBP-tdBN~\cite{2020Going}} & \multirow{3}{*}{SNN training} & \multirow{3}{*}{ResNet19} 
		& 2 & 92.34\%   \\
		&  &  &											                                  & 4 & 92.92\%   \\
		&  &  &											                                   & 6 & 93.16\%   \\
		\cline{2-6}
		& \multirow{3}{*}{TET~\cite{deng2022temporal}} & \multirow{3}{*}{SNN training} & \multirow{3}{*}{ResNet19} 
		& 2 & 94.16\%   \\
		&  &  &											                                  & 4 & 94.44\%   \\
		&  &  &											                                   & 6 & 94.50\%   \\
		\cline{2-6}
		& \multirow{3}{*}{{Real Spike~\cite{guo2022real}}} & \multirow{3}{*}{SNN training} & \multirow{3}{*}{ResNet19} 
		& 2 & {95.31\%} \\
		&  &  &											                                  & 4 & {95.51\%}  \\
		&  &  &											                                   & 6 & {96.10\%}   \\	
		\cline{2-6}	
  
		& \multirow{2}{*}{{InfLoR-SNN~\cite{guo2022reducing}}} & \multirow{2}{*}{SNN training}  & \multirow{2}{*}{ResNet20} 		                                          & 5 & {93.01\%}  \\
		&  &  &											                                  & 10 & {93.65\%}  \\
		\cline{2-6}
		& \multirow{7}{*}{\textbf{MPBN}} & \multirow{7}{*}{SNN training} & \multirow{2}{*}{ResNet19} 
		& 1 & \textbf{96.06\%}$\pm 0.10$  \\
		&  &  &											                                  & 2 & \textbf{96.47\%}$\pm 0.08$  \\
		\cline{4-6}	
		&  &  & \multirow{3}{*}{ResNet20} 		                                          & 1 & \textbf{92.22\%}$\pm 0.11$   \\
		&  &  &											                                  & 2 & \textbf{93.54\%}$\pm 0.09$   \\
		&  &  &											                                  & 4 & \textbf{94.28\%}$\pm 0.07$   \\		
		\cline{4-6}		
		&  &  &	\multirow{2}{*}{VGG16}                                                   & 2 & \textbf{93.96\%}$\pm 0.09$  \\
		&  &  &											                                 & 4 & \textbf{94.44\%}$\pm 0.08$   \\
		\bottomrule				         	
	\end{tabular}	
\end{table*}
\subsection{Loss Landscape}
We further inspect the 1D loss landscapes~\cite{li2018visualizing} of the SNNs with or without MPBN using spiking ResNet20 architecture in 2 time steps to show why the MPBN can improve accuracy and convergence speed in Fig.~\ref {loss}. 
It can be observed that the loss landscape of the SNN model with MPBN is flatter than that of the SNN without MPBN. This indicates that the MPBN makes the landscape of the corresponding optimization problem smoother~\cite{li2018visualizing}, thus making the gradients more predictive and network faster convergence.
The results here provide convincing evidence for ablation studies in Section 4.4.

\begin{table*}[htbp]
	\centering	
	\caption{Comparison with SoTA methods on CIFAR-100.}	
	\label{tab:Comparisoncifar100}	
	\begin{tabular}{lllccc}	
		\toprule
		Dataset & Method & Type & Architecture & Timestep & Accuracy \\	
		\toprule
		\multirow{27}{*}{CIFAR-100}	
		& SpikeNorm~\cite{2019Going} & ANN2SNN & ResNet20 & 2500 & 64.09\%   \\
		& RMP~\cite{2020RMP} & ANN2SNN & ResNet20 & 2048 & 67.82\%   \\
		& Hybrid-Train~\cite{2020Enabling} & Hybrid training & VGG11 & 125 & 67.90\%   \\  	
  	& IM-Loss~\cite{guo2022imloss} & SNN training & VGG16 & 5 & 70.18\%   \\ 
		\cline{2-6} 
		& \multirow{2}{*}{Joint A-SNN~\cite{guo2023joint}} & \multirow{2}{*}{SNN training} & ResNet18 & 4 & 77.39\%   \\ 
		&                                          &                                & ResNet34 & 4 & 79.76\%   \\  
		\cline{2-6}
		& \multirow{3}{*}{Dspike~\cite{li2021differentiable}} & \multirow{3}{*}{SNN training} & \multirow{3}{*}{ResNet20} 
		& 2 & 71.68\%   \\
		&  &  &											                                  & 4 & 73.35\%   \\
		&  &  &											                                   & 6 & 74.24\%   \\
		\cline{2-6}
		& \multirow{3}{*}{TET~\cite{deng2022temporal}} & \multirow{3}{*}{SNN training} & \multirow{3}{*}{ResNet19} 
		& 2 & 72.87\%   \\
		&  &  &											                                  & 4 & 74.47\%   \\
		&  &  &											                                   & 6 & 74.72\%   \\
		\cline{2-6}
		& \multirow{2}{*}{RecDis-SNN~\cite{Guo_2022_CVPR}} & \multirow{2}{*}{SNN training} & ResNet19 & 4 & 74.10\%   \\ 
		&                                          &                                & VGG16 & 5 & 69.88\%   \\  
		\cline{2-6}
		& \multirow{2}{*}{{InfLoR-SNN~\cite{guo2022reducing}}} & \multirow{2}{*}{SNN training}  & ResNet20 		                                          & 5 & {71.19\%}  \\
		&  &  &											                                 VGG16 & 5 & {71.56\%}  \\
  \cline{2-6}
		& \multirow{2}{*}{Real Spike~\cite{guo2022real}} & \multirow{2}{*}{SNN training} & ResNet20 & 5 & 66.60\%   \\ 
		&                                          &                                & VGG16 & 5 & 70.62\%   \\  
		\cline{2-6}
		& \multirow{2}{*}{GLIF~\cite{yao2022glif}} & \multirow{2}{*}{SNN training} & \multirow{2}{*}{ResNet19} & 2 & 75.48\%   \\ 
		&                                          &                                &   & 4 & 77.05\%   \\  
		\cline{2-6}

		& \multirow{3}{*}{TEBN~\cite{duan2022temporal}} & \multirow{3}{*}{SNN training} & \multirow{3}{*}{ResNet19} 
		& 2 & 75.86\%   \\
		&  &  &											                                  & 4 & 76.13\%   \\
		&  &  &											                                   & 6 & 76.41\%   \\
  \cline{2-6}	
  		& \multirow{5}{*}{\textbf{MPBN} } & \multirow{5}{*}{SNN training} & {VGG16} 
		& 4 & \textbf{74.74\%}$\pm 0.11$   \\
		\cline{4-6}	
		&  &  &	\multirow{2}{*}{ResNet19}                                                       & 1 & \textbf{78.71\%}$\pm 0.10$   \\
		&  &  &											                                     & 2 & \textbf{79.51\%}$\pm 0.07$   \\
		\cline{4-6}	
		&  &  &	\multirow{2}{*}{ResNet20}                                                    & 2 & \textbf{70.79\%}$\pm 0.08$  \\
		&  &  &											                                     & 4 & \textbf{72.30\%}$\pm 0.08$  \\
		\bottomrule				         	
	\end{tabular}	
\end{table*}
\section{Experiments}

In this section, abundant experiments were conducted to verify the effectiveness of the MPBN using widely-used spiking ResNet20~\cite{2020DIET,2019Going}, VGG16~\cite{2020DIET}, ResNet18~\cite{2021Deep},  ResNet19~\cite{2020Going}, and ResNet34~\cite{2021Deep} on both static datasets  including CIFAR-10~\cite{CIFAR-10}, CIFAR-100~\cite{CIFAR-10}, and ImageNet~\cite{2009ImageNet}, and one neuromorphic dataset, CIFAR10-DVS~\cite{2017CIFAR10}. 
The specific introduction for these datasets has be detailed in many recent works~\cite{2020Going,2020DIET,guo2022real,li2021differentiable}. Here, we mainly introduce these hyper-parameters and data preprocessing  in detail. We used the widely adopted LIF neuron in our SNN models as other works about direct training methods~\cite{2020DIET,2019Going}. These hyper-parameters for LIF neuron about the initial firing threshold $V_{\rm th}$ and the membrane potential decaying constant $\tau_{\rm decay}$ are $0.5$ and $0.25$ respectively. 
For static image datasets, since encoding the 8-bit RGB images into 1-bit spikes will lose too much information, we use an ANN-like convolutional layer and a LIF layer to encode the images to spikes for
all the rest of the layers, as in recent works~\cite{2020Going,2020DIET,guo2022real,li2021differentiable}. 

\subsection{Comparison with SoTA Methods}

\begin{table*}[htbp]
	\centering	
	\caption{Comparison with SoTA methods on ImageNet.}	
	\label{tab:Comparisonimage}	
	\begin{tabular}{lllccc}	
		\toprule
		Dataset & Method & Type & Architecture & Timestep &  Accuracy \\	
		\toprule
       \multirow{9}{*}{ImageNet}
		&STBP-tdBN~\cite{2020Going} &  SNN training & ResNet34 & 6 & 63.72\%   \\ 
		&TET~\cite{deng2022temporal} &  SNN training & ResNet34 & 6 & 64.79\%   \\
		&MS-ResNet~\cite{hu2021advancing} &  SNN training & ResNet18 & 6 & 63.10\%   \\
  	&OTTT~\cite{xiao2022online} &  SNN training & ResNet34 & 6 & 63.10\%   \\
		&\multirow{1}{*}{Real Spike~\cite{guo2022real}} & \multirow{1}{*}{SNN training} & {ResNet18} & 4  & {63.68\%}   \\
		\cline{2-6}		
		&\multirow{2}{*}{SEW ResNet~\cite{2021Deep}} & \multirow{2}{*}{SNN training} & {ResNet18} & 4  & {63.18\%}   \\
		& &  &											                             {ResNet34} & 4  & {67.04\%}   \\ 
		\cline{2-6}
		&\multirow{2}{*}{\textbf{MPBN} } & \multirow{2}{*}{SNN training} & {ResNet18} & 4 & \textbf{63.14\%}$\pm 0.08$   \\
		& &  &											                             {ResNet34} & 4& \textbf{64.71\%}$\pm 0.09$   \\
		\bottomrule				         	
	\end{tabular}	
\end{table*}

\begin{table*}[htbp]
	\centering	
	\caption{Comparison with SoTA methods on CIFAR10-DVS.}	
	\label{tab:Comparisondvs}	
	\begin{tabular}{lllccc}	
		\toprule
		 Dataset & Method & Type & Architecture & Timestep & Accuracy \\	
		\toprule
  \multirow{9}{*}{CIFAR10-DVS}
		 &Rollout~\cite{2020Efficient} & Rollout & DenseNet & 10 & 66.80\%   \\
		 &LIAF-Net~\cite{LIAF-Net} & Conv3D & LIAF-Net & 10 &  71.70\%   \\
		 &LIAF-Net~\cite{LIAF-Net} & LIAF & LIAF-Net & 10 & 70.40\%   \\
		&STBP-tdBN~\cite{2020Going} & SNN training & ResNet19 & 10 & 67.80\%   \\ 
		&RecDis-SNN~\cite{Guo_2022_CVPR} & SNN training & ResNet19 & 10 & 72.42\%   \\ 
		\cline{2-6}
		& \multirow{2}{*}{Real Spike~\cite{guo2022real}} & \multirow{2}{*}{SNN training} & {ResNet19} 
		& 10 & {72.85\%}   \\
		&  &  &											                 {ResNet20} & 10 & {78.00\%}   \\
		\cline{2-6}
		& \multirow{2}{*}{\textbf{MPBN}} & \multirow{2}{*}{SNN training} & {ResNet19} 
		& 10 & \textbf{74.40\%}$\pm 0.20$   \\
		&  &  &											                 {ResNet20} & 10 & \textbf{78.70\%}$\pm 0.10$   \\
		\bottomrule				         	
	\end{tabular}	
\end{table*}

\textbf{CIFAR-10.} 
On CIFAR-10, we trained our SNN model using the SGD optimizer with 0.9 momentum. The initial learning rate is 0.1 and decays to 0 in cosine form. The total training time is 400 epochs.
To fairly compare with these recent SoTA methods~\cite{li2021differentiable,Guo_2022_CVPR,guo2022reducing}, we also adopt data normalization, random horizontal flipping, cropping,  and cutout~\cite{devries2017improved} for data augmentation. We run three times for each experiment to report the ``mean ± std'' in Tab.~\ref{tab:Comparisoncifar}. 
It can be seen that our models can outperform other methods over all these chosen widely adopted architectures with fewer time steps. 
For example, The accuracy of spiking ResNet19 trained with MPBN with only 1 time step can be up to 96.06\%, while the Real Spike ~\cite{guo2022real} needs 6 time steps to reach a comparable result and the RecDis-SNN~\cite{2020Temporal} even still underperforms 0.51\% with 6 time steps.
this superiority can also be observed in the results regarding the spiking ResNet20 and VGG16.

\textbf{CIFAR-100.} 
For CIFAR-100, we adopted the same settings as in CIFAR-10. 
The proposed MPBN also performs well on CIFAR-100. It can be seen that our method gets the best accuracy over all these networks even with fewer time steps. 
For instants, the ResNet19 trained with MPBN can achieve 78.71\% top-1 accuracy with only 1 time step, which outperforms other SoTA methods such as TET, GLIF, TEBN, and RecDis-SNN even with  4 or 6 time steps about 1.66\%-3.99\%  .

\textbf{ImageNet.}
On ImageNet, we used standard data normalization, random horizontal flipping, and cropping for data augmentation and trained the networks for 320 epochs as in~\cite{2021Deep}. The optimizer setting also keeps the same with CIFAR datasets. 
The results for ImageNet are presented in Tab.~\ref{tab:Comparisonimage}.
It can be seen that the accuracy of our method is  better than that of these recent SoTA methods, only relatively smaller compared with SEW ResNet~\cite{2021Deep} for spiking ResNet34. 
However, SEW ResNet is not a typical SNN model. It adopts the activation before addition form-based ResNet and its blocks  will fire positive integer spikes. In this way, the event-driven and multiplication-addition transform advantages of SNNs will be lost. While we adopt the original ResNet, which fires standard binary spikes.

\textbf{CIFAR10-DVS.}
We also adopted the neuromorphic dataset, CIFAR10-DVS in the paper to verify the effectiveness of the MPBN. We also split the dataset into 9K training images and 1K  test images, and resize them to $48\times 48$ for data augmentation as in~\cite{2018Direct,guo2022real}. The learning rate is 0.01 and other settings are the same as CIFAR-10. It can be seen that the MPBN also shows superiority in this dataset. 

\subsection{ Extension of the MPBN}
In CNNs, the most widely used BN is channel-wised. This is because element-wised BNs are very time-consuming and can not be folded into weights, otherwise, the channel-wised weight-sharing mechanism will be destroyed. However, the MPBN adopts the firing threshold-folded manner and the firing threshold need not keep the same along the channels, therefore, MPBN can use the element-wised form freely. In this way, $V_{\rm th}$ will be transformed to element-wised ones as follows,
\begin{equation}
 (\bm{\tilde{V}}_{\rm th})_{i,j,k} = \frac{(V_{\rm th} - \bm{\beta}_{i,j,k}){\sqrt{\bm{\sigma}_{i,j,k}^2}}}{\bm{\lambda}_{i,j,k}} + \bm{\mu}_{i,j,k},
    \label{eq_lif_bn3}
\end{equation}
where$(\bm{\tilde{V}}_{\rm th})_{i,j,k}$ is the transformed firing threshold of the neuron comes from $i$-th channel in the spatial position $(j,k)$. To investigate the performance of the element-wised MPBN, here we also provide a comparison of the vanilla MPBN and its extension. The results of top-1 accuracy of the spiking ResNet20 with 4 time steps on CIFAR datasets are shown in Tab.~\ref {ext}. Though the two versions all perform well, the element-wised MPBN is relatively better than the channel-wised MPBN. This may be because element-wised MPBN can learn more firing threshold values, which means a richer representation ability for SNNs.

\begin{table}[]	
	\centering	
	\caption{Comparison with learnable firing threshold methods.}	
	\label{ext}	
	\begin{tabular}{clcc}	
		\toprule
		Dataset & Method & Time step & Accuracy \\	
		\toprule
		\multirow{2}{*}{CIFAR-10}		
        & channel-wised   & 4 & 94.28\%   \\
		& element-wised   & 4 & 94.42\%   \\
        \cline{1-4}
        \multirow{2}{*}{CIFAR-100}		
        & channel-wised  & 4 & 72.30\%   \\
		& element-wised   & 4 & 72.49\%   \\
		\bottomrule 	
	\end{tabular}	
\end{table}

\section{Conclusion}
In the paper, we advocated adding the MPBN before the firing function to regulate the disturbed data flow again. 
We also provided a training-inference-decoupled re-parameterization technique to fold the trained MPBN into the firing threshold to eliminate the extra time burden induced by MPBN in the inference time.
Furthermore, the channel-wised and element-wised MPBN in different granularities were explored. 
Extensive experiments verified that the proposed MPBN can consistently achieve good performance.

\section*{Acknowledgment}
This work is supported by grants from the National Natural Science Foundation of China under
contracts No.12202412 and No.12202413.

{\small
\bibliographystyle{ieee_fullname}
\bibliography{egbib}
}

\end{document}